\documentclass[final]{cvpr}

\usepackage{times}
\usepackage{epsfig}
\usepackage{graphicx}
\usepackage{amsmath}
\usepackage{amssymb}
\usepackage{float}
\usepackage{textgreek}
\usepackage{enumitem}
\usepackage[utf8]{inputenc}
\usepackage[table]{xcolor}
\usepackage{booktabs, multirow}
\usepackage{adjustbox}

\usepackage{hyperref}
\hypersetup{pagebackref=true,breaklinks=true,colorlinks,bookmarks=false}

\def\cvprPaperID{****} 
\def\confYear{CVPR 2021}

\begin{document}

\title{Investigating the Challenges of Class Imbalance and Scale Variation in Object Detection in Aerial Images}

\author{Ahmed Adel Gomaa Elhagry\\
\tt\small Ahmed.Elhagry@mbzuai.ac.ae
\and
Mohamed Eltag Salih Saeed\\
{\tt\small Mohamed.Saeed@mbzuai.ac.ae}
}

\maketitle

\begin{abstract}
While object detection is a common problem in computer vision, it is even more challenging when dealing with aerial satellite images. The variety in object scales and orientations can make them difficult to identify. In addition, there can be large amounts of densely packed small objects such as cars. In this project, we propose a few changes to the Faster-RCNN architecture. First, we experiment with different backbones to extract better features. We also modify the data augmentations and generated anchor sizes for region proposals in order to better handle small objects. Finally, we investigate the effects of different loss functions. Our proposed design achieves an improvement of 4.7 mAP over the baseline which used a vanilla Faster R-CNN with a ResNet-101 FPN backbone.
\end{abstract}

\section{Introduction}

\subsection{Problem Statement}
The problem of object detection involves a combination of two tasks: classification and localization; identifying object classes and drawing a bounding box around them. Remote sensing applications require accurate object detection from aerial images. Such images are more difficult to deal with due to the variety of objects that can be present, in addition to significant variations in scales and orientations. For this project, we aim to tackle this problem using the Instance Segmentation in Aerial Images (iSAID) dataset \cite{isaid}. 

\begin{figure}[h]
    \centering
    \includegraphics[width=\linewidth, keepaspectratio=true]{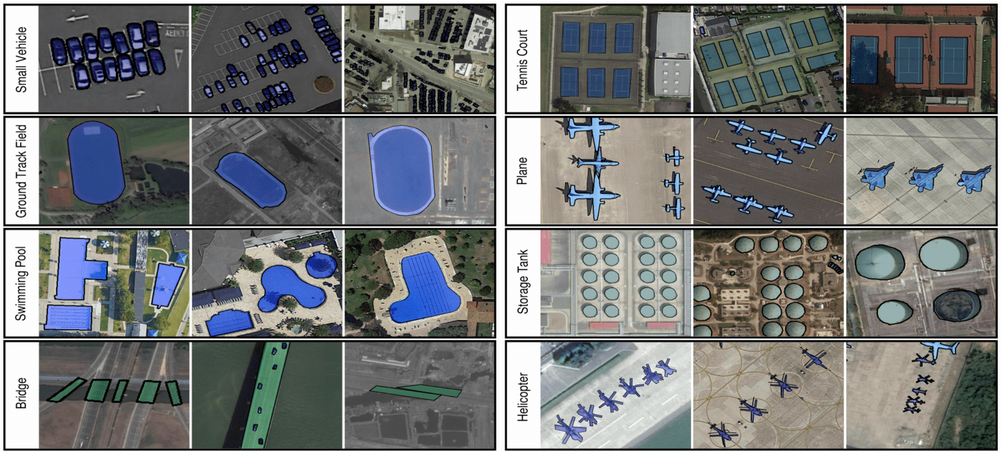}
    \caption{Images of objects from a few classes in the iSAID dataset. Variations in scale can be seen among the classes, in addition to the densely packed small objects in the small vehicle class. \cite{isaid}}
    \label{fig:scaleVariations}
\end{figure}

\subsection{Datasets}

\subsubsection{iSAID}
\begin{figure}[h]
    \centering
    \includegraphics[width=\linewidth, keepaspectratio=true]{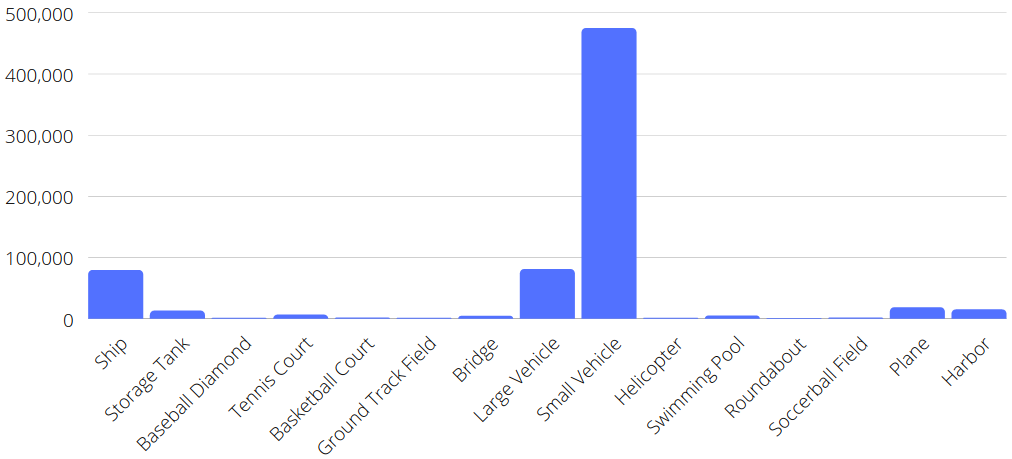}
    \caption{Imbalanced and uneven distribution of objects of the iSAID dataset \cite{isaid}}
    \label{fig:classImbalance}
\end{figure}

The Instance Segmentation in Aerial Images Dataset (iSAID) \cite{isaid} is a dataset containing 2,806 high resolution satellite images with dense annotations of 655,451 instances of various objects such as ships, harbors, helicopters, bridges etc (Fig. \ref{fig:scaleVariations}). There is a total of 15 classes within the dataset. The images in the iSAID dataset originate from the DOTA dataset \cite{dota}, but the annotations are significantly improved upon. Instance segmentation and object detection are the two tasks the dataset was designed for. The focus of this project is the detection task. However, there are challenges when it comes to this. The first challenge is the class imbalance within the dataset. In addition, most of the instances are small vehicles as shown in Fig. \ref{fig:classImbalance}. Their relative size can also be seen in Fig. \ref{fig:scaleVariations} which shows examples of a few of the object classes in the dataset.

The second challenge to deal with is the small objects and scale variations, given that the most occurring class is the smallest one in size, which is the small vehicles. There are also scale variations within the same class. Relative sizes of objects and variations exist as shown in Fig. \ref{fig:scaleVariations}, where for example an image can have about a hundred small vehicles in it, whereas one ground track field can take up a whole image.

\subsubsection{COCO}
The Common Objects in Context (COCO) dataset is an object detection, segmentation and captioning dataset. It contains over 200,000 labelled images with 1.5 million instances of objects from 80 different categories.

\section{Related Work}

\subsection{Faster R-CNN}

\begin{figure}[h]
    \centering
    \includegraphics[width=\linewidth, keepaspectratio=true]{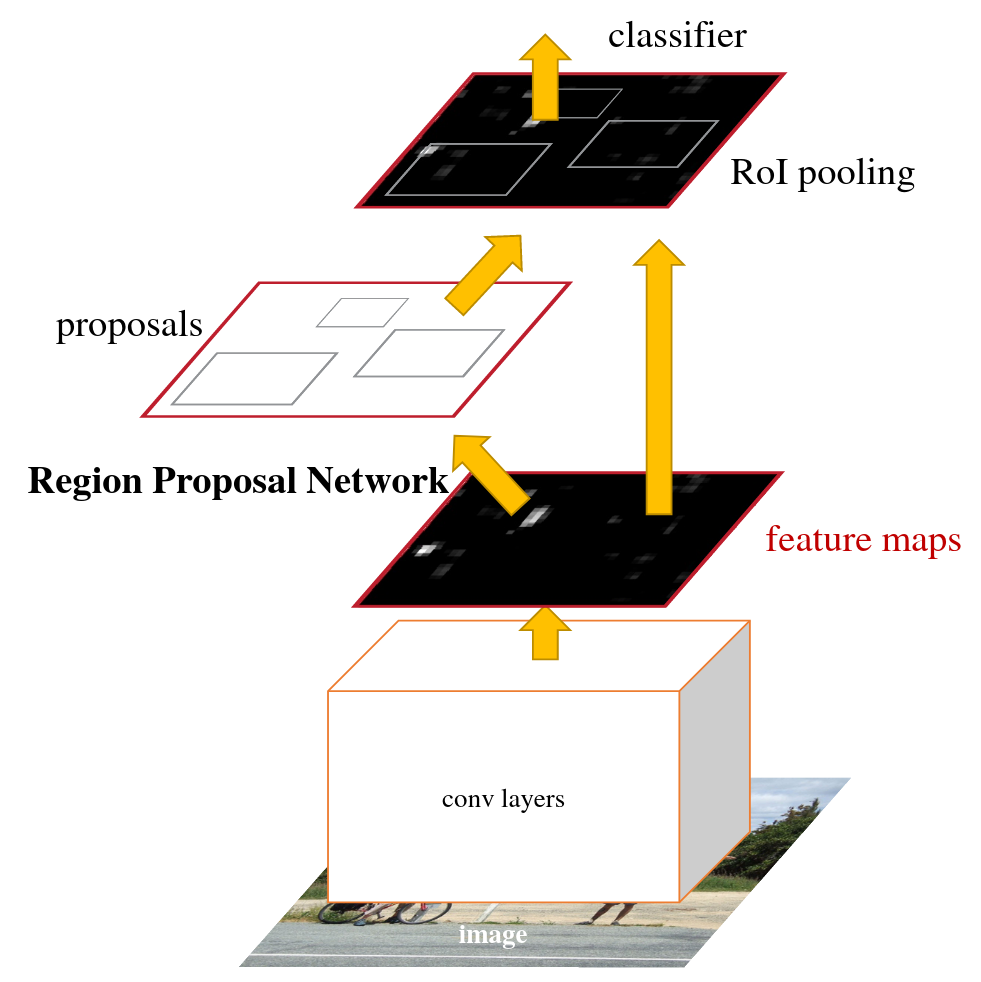}
    \caption{The original Faster R-CNN architecture \cite{ren2015faster}}
    \label{fig:faster}
\end{figure}

Faster R-CNN \cite{he2017mask} is an architecture technique that is commonly used in object detection tasks. It is divided into two phases (Figure \ref{fig:faster}). First, it creates proposals based on the inserted image for locations where an object could be present. Second, based on the first stage's proposal, it predicts the object's class, refines the predicted bounding box for the object. Both stages share a common backbone, which is a convolutional neural network that extracts features from the raw images. Typically, the backbones are based on ResNet \cite{resnet}.

In the first stage, the Region Proposal Network (RPN) searches the backbone's feature maps and suggests where objects could be found. Anchors are used for this, which are boxes of various predefined scales and aspect ratios, generated for every point in the extracted feature map. Individual anchors are then classified for objectness (whether they contain an object or not). In addition, the bounding box coordinates are refined by the network's regression branch to improve the localization.

The second stage is the object detection stage. For feature extraction, this stage shares the same backbone as the first stage. It then uses the RPN's proposals to pool the extracted features from the backbone. This part is called the ROI pooling layer and it splits every region proposal into multiple sub-areas and does max pooling on them to generate a fixed size output. This is then fed to 2 fully connected layers and then to a classification branch and a regression branch to refine the bounding box.

\subsection{Backbone Networks}

\subsubsection{ResNet}
Residual networks, also known as ResNets \cite{resnet}, are a class of CNNs that make use of skip connections to pass outputs from earlier layers in the network to later layers. This design was found to be very useful in training deeper networks than what was possible at the time. The authors came up with multiple designs of varying depths based on this idea. The basic building block of the network is the bottleneck (Fig. \ref{fig:ResNext} (left)), consisting of two 1x1 convolution layers with 3x3 convolutions in between and a residual skip connection to add the bottleneck's input to the output of the convolutions.

\subsubsection{ResNeXt}
\begin{figure}[h]
    \centering
    \includegraphics[width=\linewidth, keepaspectratio=true]{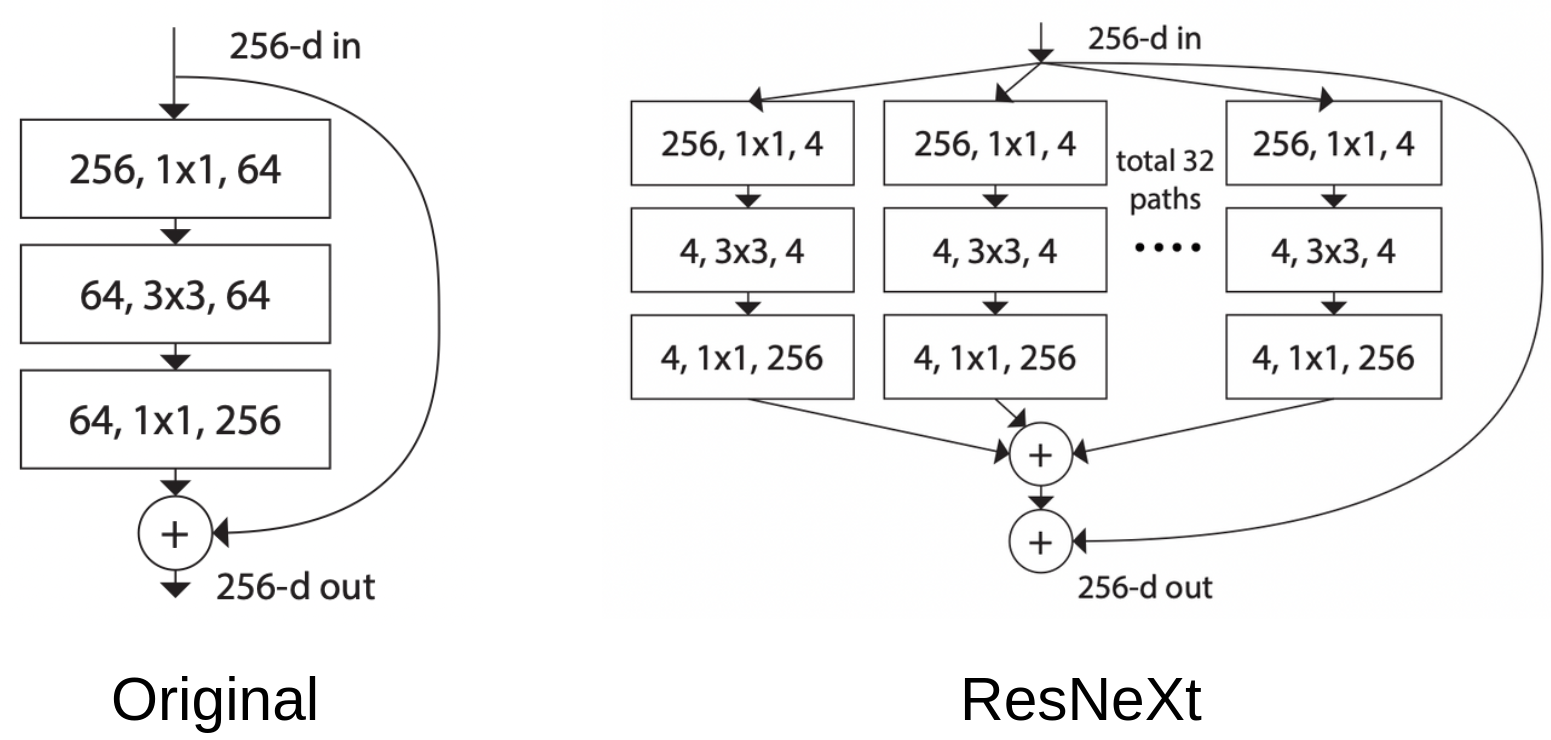}
    \caption{\textbf{Left:} The original ResNet bottleneck \cite{resnet} which is the network's basic building block. \textbf{Right:} The modified ResNext bottleneck \cite{resnext}}
    \label{fig:ResNext}
\end{figure}

ResNeXt \cite{resnext} is a variation of the original ResNet architecture that splits the bottleneck convolutions into groups of 4, making the computations more efficient. Figure \ref{fig:ResNext} below compares a ResNet bottleneck with the ResNeXt bottleneck.

\subsubsection{Res2Net}
\begin{figure}[h]
    \centering
    \includegraphics[width=\linewidth, keepaspectratio=true]{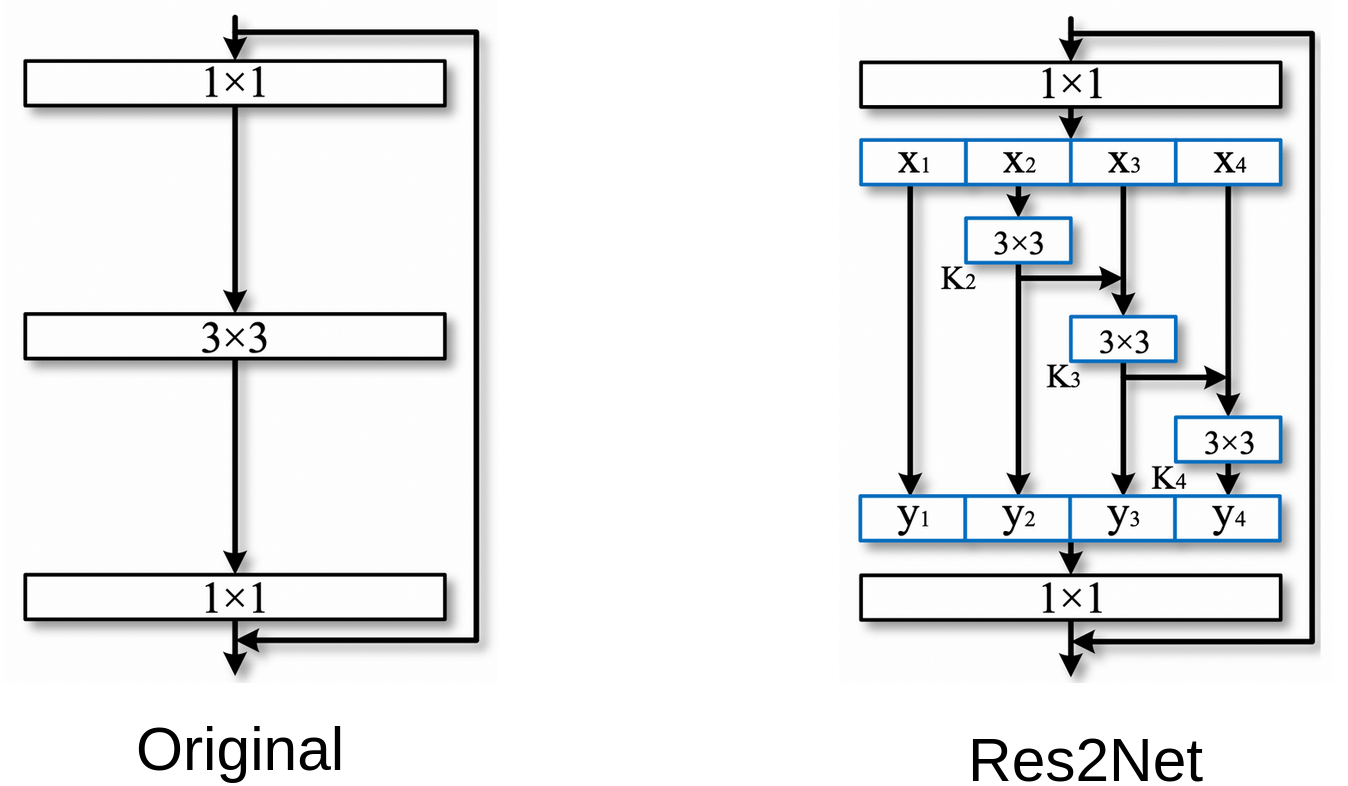}
    \caption{Res2Net Architecture \cite{res2net}}
    \label{fig:Res2Net}
\end{figure}

Res2Net follows the same pipeline as the original ResNet but modifies the bottleneck block of the network as shown in Fig. \ref{fig:Res2Net}. Rather than having a single 3x3 conv, there are three of them. Four pathways are consequently available for the input coming in from the first 1x1 conv. The first one does not go through any 3x3 conv, the second goes through one of them, the third goes through two and so on. This essentially allows multi-scale feature extraction within the bottleneck itself.

\subsubsection{Feature Pyramid Networks}

\begin{figure}[h]
    \centering
    \includegraphics[width=\linewidth, keepaspectratio=true]{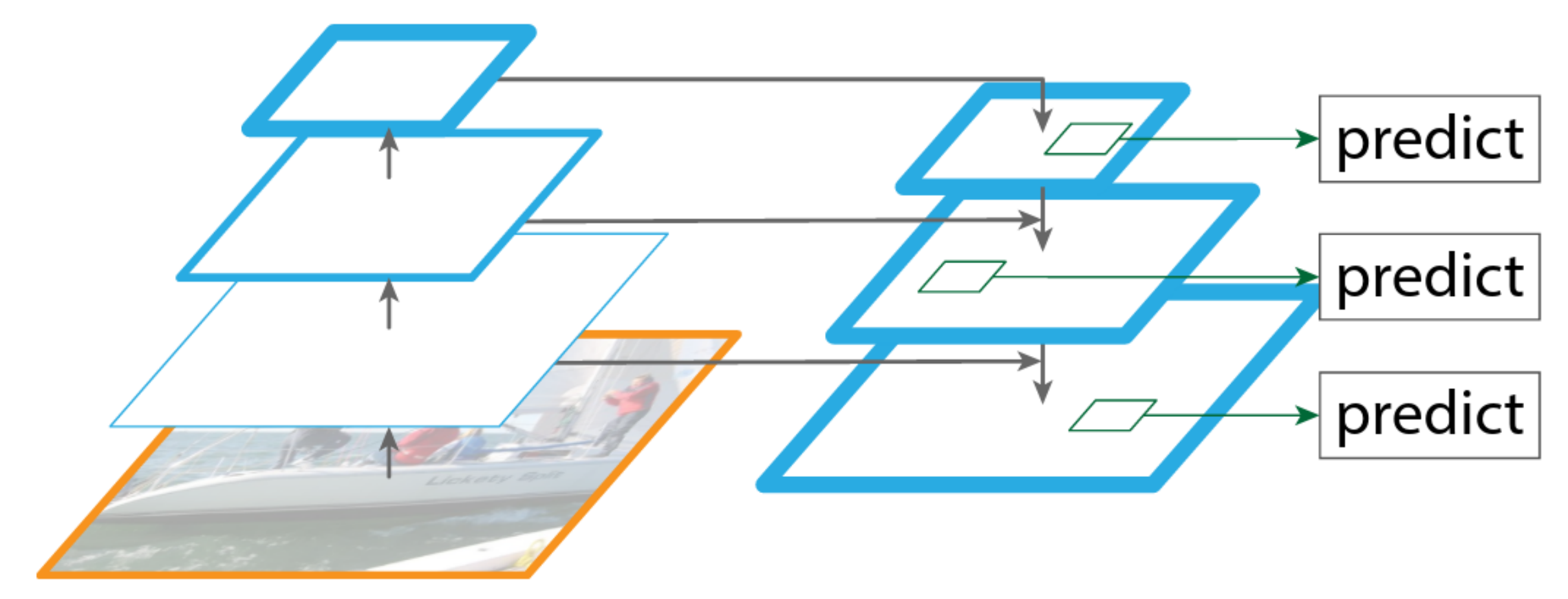}
    \caption{Outline of the feature pyramid architecture with connected bottom-up and top-down pathways \cite{fpn}}
    \label{fig:fpn}
\end{figure}

A feature pyramid network (FPN) \cite{fpn} captures features at multiple scales using a bottom-up pathway, a top-down pathway and lateral connections between corresponding levels (Fig. \ref{fig:fpn}. A feature extractor such can be used as a bottom-up pathway, usually a ResNet. The feature pyramid map generated by the top-bottom pathway is the same size of the feature pyramid map generated by the bottom-up method. Convolution and concatenating operations are performed between two matching levels of the two paths to form the lateral connections.

\subsubsection{Dilated \& Depthwise Separable Convolutions}

Depthwise separable convolutions consist of a depthwise convolution, in which an independent kernel is used for each channel, followed by a pointwise (1x1) convolution with a kernel that has a depth equal to the number of channels. This greatly decreases the number of parameters and hence the computational cost, at the expense of a reduction in accuracy \cite{kim2020lightweight}. Meanwhile, dilated convolutions can also be done to increase the receptive field without increasing the computational cost much. This is done by applying the kernel to an input with gaps that are created by skipping some pixels. Relative to standard convolutions, dilated convolutions have the same or lower number of parameters, depending on the dilation scale, but result in a wider receptive field \cite{kim2020lightweight}. These can both be integrated into a ResNet bottleneck to increase accuracy or efficiency.
\subsection{Loss functions}

\subsubsection{Focal Loss}

\begin{figure}[h]
    \centering
    \includegraphics[width=1.0\linewidth, keepaspectratio=true]{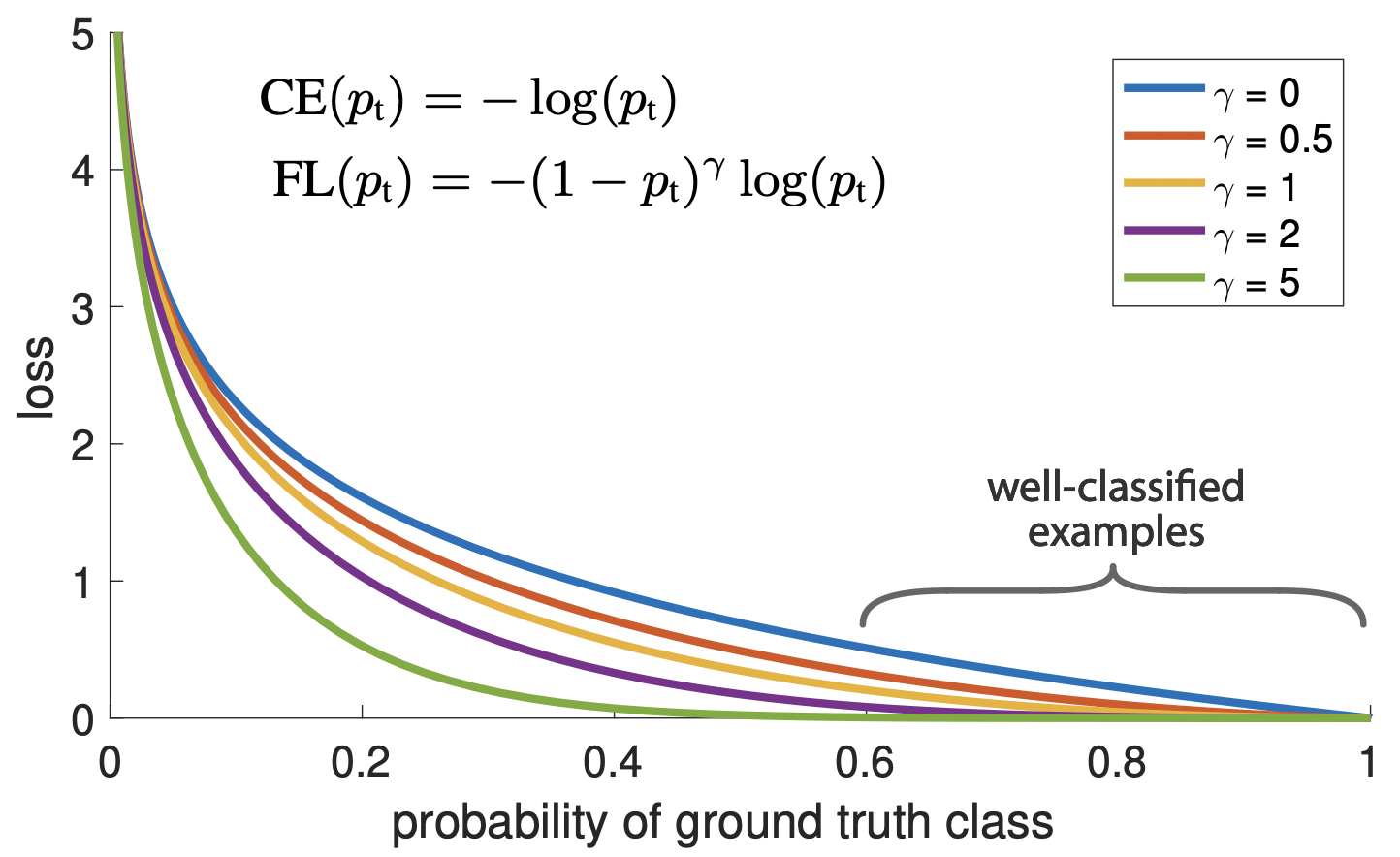}
    \caption{The effect of the focal loss function \cite{lin2017focal}}
    \label{fig:focal}
\end{figure}

Previous work in object detection by \cite{lin2017focal} found that a major issue in dense object detectors is the imbalance between foreground and background classes. Apart from the network architecture that they designed, RetinaNet, they also proposed a solution to this problem in the form of a new loss function that they call focal loss. It is a modification on cross entropy loss in order to reduce the weight given to the loss when it comes to examples that the model can classify well. In essence, the loss 'focuses' on the misclassifed examples. Fig. \ref{fig:focal} shows an illustration of the effect of using focal loss.

\section{Methods}

\textbf{Detectron2} \cite{wu2019detectron2}, which is a library offering state-of-the-art algorithms for both segmentation and detection tasks, was used to run the experiments. Our main goal is to study the effect of different modifications on the backbone, loss functions, augmentations, anchor sizes and angles during fine-tuning on iSAID dataset for the task of object detection. All ResNet and ResNext pretrained backbones that are not modified use weights from COCO pretraining.

\subsection{Baseline}

\begin{figure}[h]
    \centering
    \includegraphics[width=\linewidth, keepaspectratio=true]{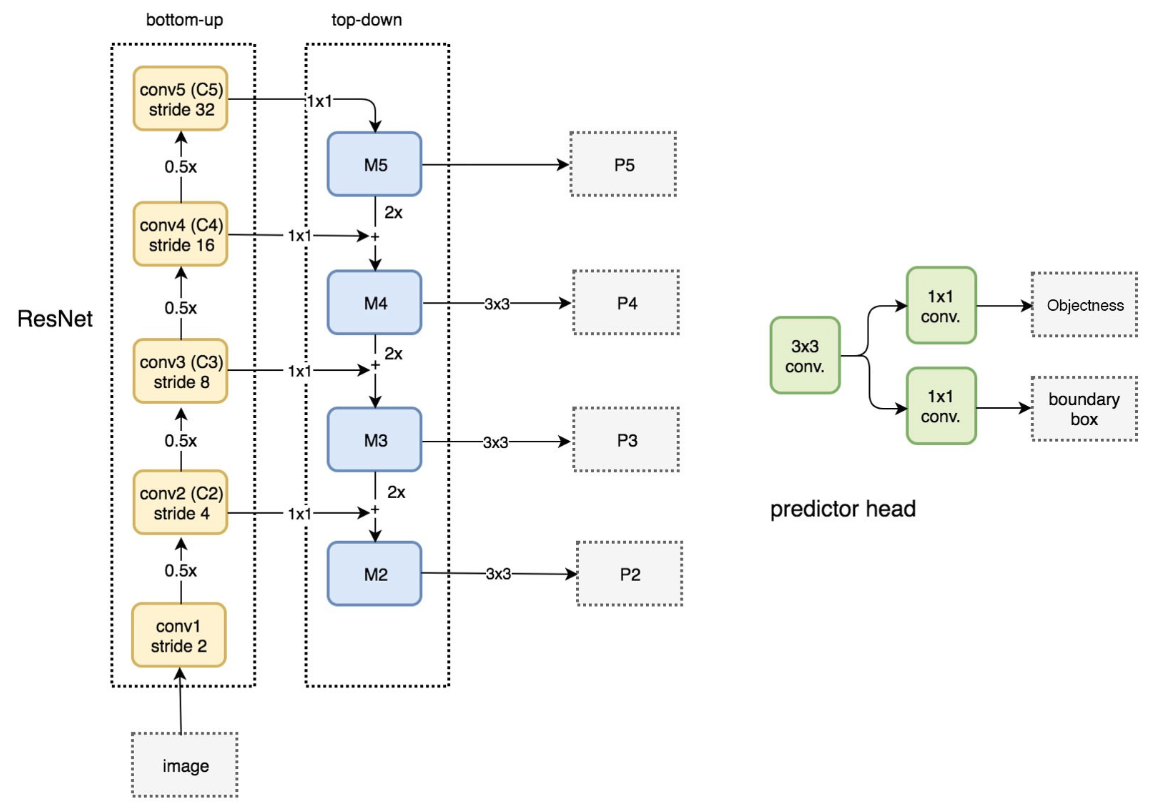}
    \caption{Faster R-CNN with a ResNet FPN backbone \cite{fasterRCNN}}
    \label{fig:fasterRCNN}
\end{figure}

Our baseline is a Faster R-CNN with a ResNet-101 FPN backbone \cite{fasterRCNNPaper}, as shown in Fig. \ref{fig:fasterRCNN}. We tried to improve upon this by modifying the network architecture; mostly working on the ResNet architecture. We also modified the loss functions, tried a variety of augmentations, and also tried to reach the best anchor sizes and angles as a prior to our problem. 

\subsection{ResNet Tweaks}

\subsubsection{First 7x7 convolution}

\begin{figure}[h]
    \centering
    \includegraphics[width=\linewidth, keepaspectratio=true]{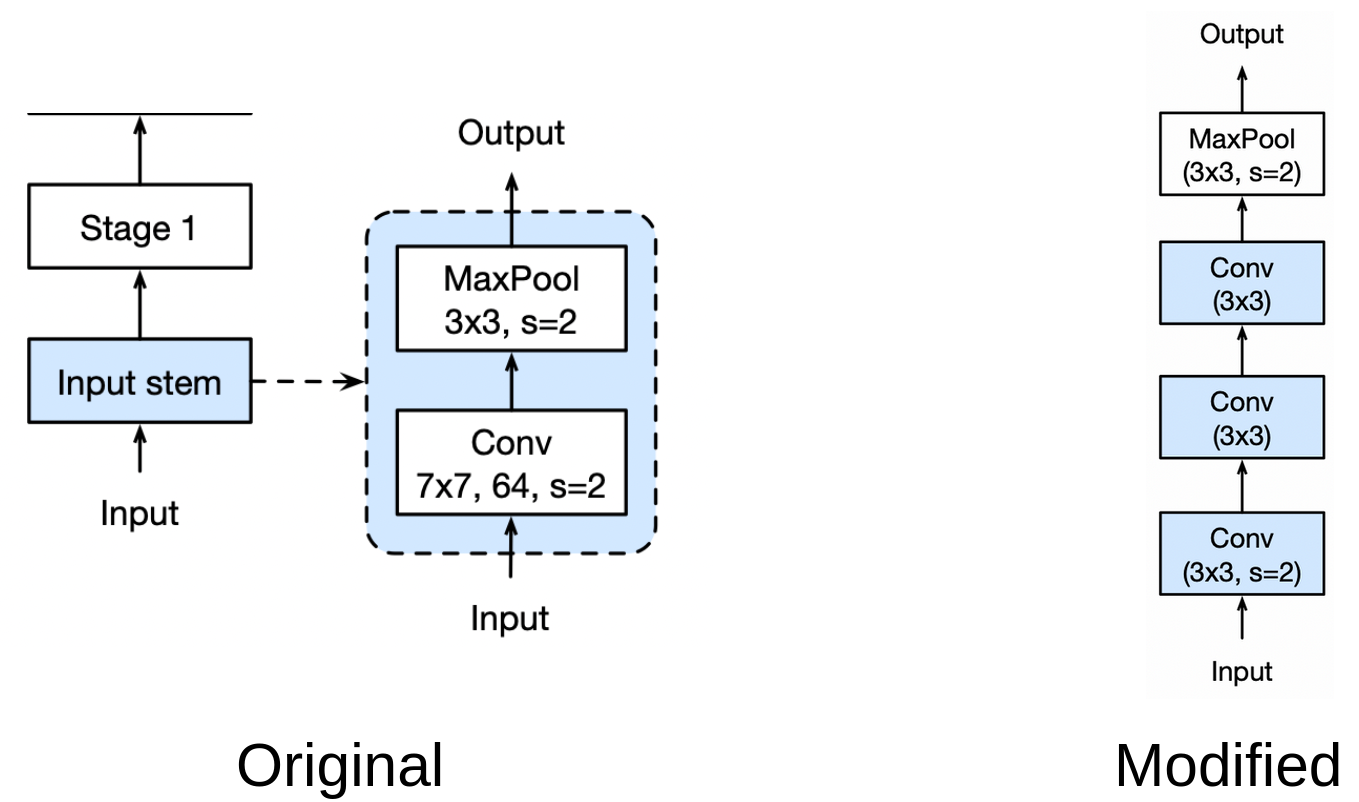}
    \caption{Replacement of the first 7x7 convolution with 3 3x3 convolutions \cite{bagsOfTricks}}
    \label{fig:1stTweak}
\end{figure}

In terms of network modifications, the first tweak we made was for efficiency. As mentioned in \cite{bagsOfTricks}, we changed the first convolution in ResNet from a 7x7 convolutions to 3 3x3 convolutions, as shown in Fig. \ref{fig:1stTweak}. This gives us fewer parameters, and it makes the network slightly faster.

\subsubsection{Stride in 1x1 convolution}

\begin{figure}[h]
    \centering
    \includegraphics[width=\linewidth, keepaspectratio=true]{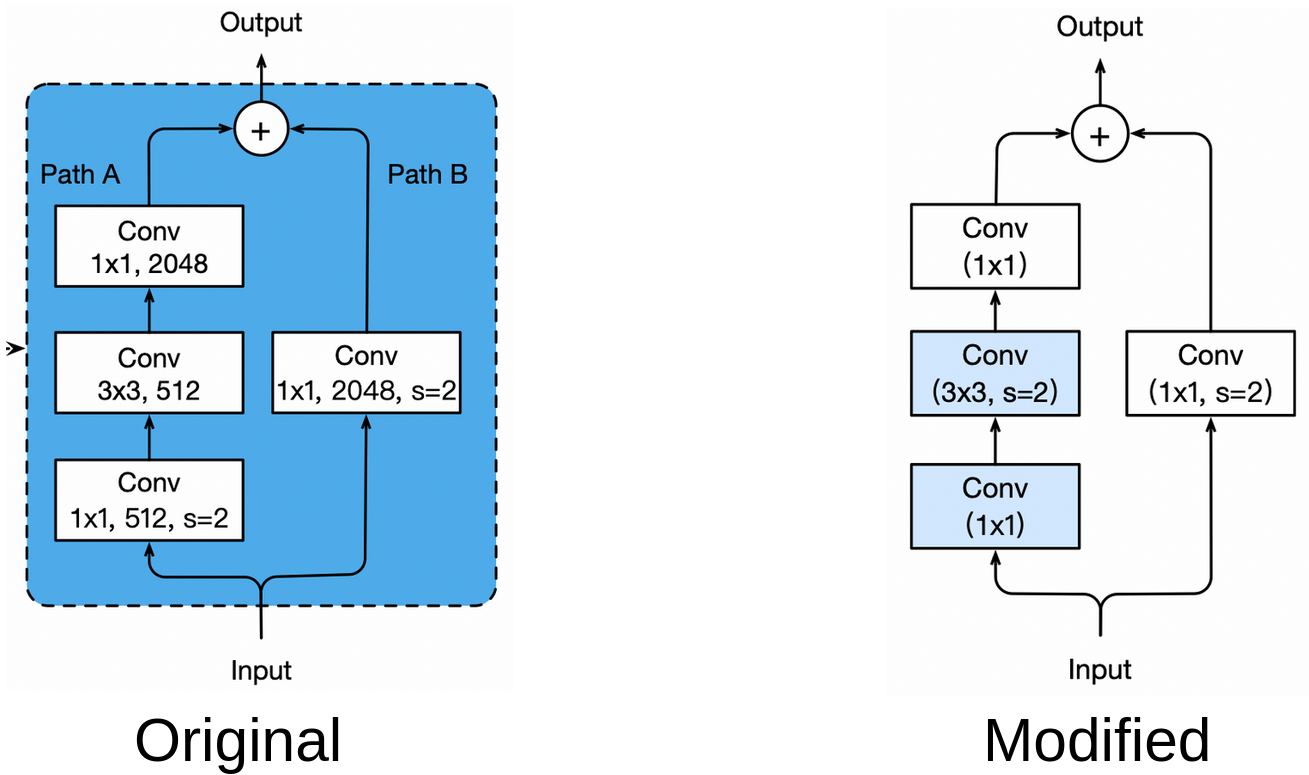}
    \caption{From a ResNet 4th stage stride 2 in the 1x1 convolution to the 3x3 convolution \cite{bagsOfTricks}}
    \label{fig:2ndTweak}
\end{figure}

The second network tweak is in the fourth stage of ResNet in the down-sampling. The original architecture starts with a stride 2 in the 1x1 convolution. We move this to the 3x3 convolution to avoid losing information in the first convolution, as shown in Fig. \ref{fig:2ndTweak}. Because of it is kept at the stride in the beginning, the gaps will cause the information loss \cite{bagsOfTricks}.

\subsubsection{Multidilated Depthwise Separable Convolutions}
\begin{figure}[h]
    \centering
    \includegraphics[width=\linewidth, keepaspectratio=true]{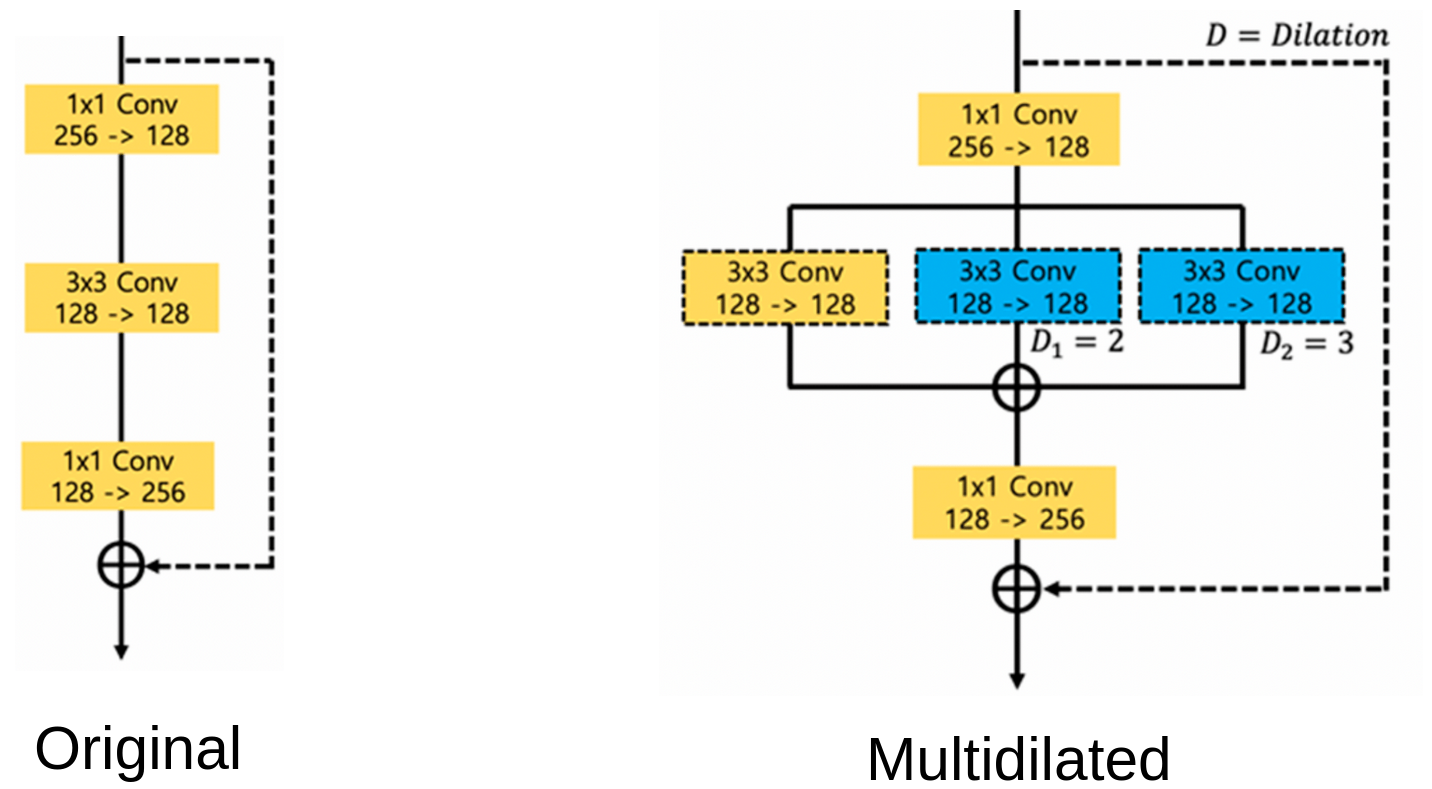}
    \caption{ResNet bottleneck with multidilated depthwise separable convolutions \cite{kim2020lightweight}}
    \label{fig:multidilationComparison}
\end{figure}

The aim of using multidilated depthwise separable convolutions is to encode features of different scales within the network. Dilation effectively achieves this. As shown in Fig. \ref{fig:multidilationComparison}, we overall have our normal 3x3 convolution and a 3x3 convolution with a dilation of 2 and a 3x3 convolution with dilation of 3. Therefore, we are looking at different receptive fields within the same level, and then we concatenate these features.  

\subsection{Backbone Changes}

\subsubsection{ResNet-50}
Here we replace the ResNet-101 backbone with a ResNet-50 backbone with the goal of potentially achieving similar performance more efficiently. This is combined with other changes since the base ResNet-50 does not outperform the ResNet-101 on its own.

\subsubsection{ResNeXt}
With this backbone, convolutions are grouped into groups of four, so the hidden layers are of four input channels and four output channels. This is replacing the original 64x64 in the original model \cite{resnext}, as shown in Fig. \ref{fig:ResNext}. The aim is to have the same number of parameters, but with more features. This significantly reflects in the efficiency of the overall architecture. 

\subsubsection{Res2Net}
Having the same pipeline of ResNet, Res2Net follows the same intuition but in series instead of parallel, as shown in Fig. \ref{fig:Res2Net}. So we have the original input $y_1$, and we have $x_2$ going through one convolution giving us $y_2$. Same for $x_3$ going through two convolutions giving us $y_3$, and $x_4$ going through three convolutions giving us $y_4$. This helps at representing the features at multiple scales, which is crucial for our task \cite{res2net}.

\subsection{Anchor Sizes}

Initially, we simply used smaller anchors of sizes (16, 32, 64, 128, 256) and (8, 16, 32, 64, 128) instead of the original configuration. However, inspired by YOLO9000 \cite{redmon2017yolo9000}, we decided to try precomputed anchor sizes by clustering the ground truth bounding boxes into \textit{k} clusters, based on their dimensions. The distance used for clustering is the intersection over union (IoU) distance rather than Euclidean distance. For every \textit{k}, we also compute the mean IOU between the cluster centres and the actual data points. Obviously, larger values of \textit{k} would represent the dataset better, but we had to find a good balance between \textit{k} and the mean IoU since increasing \textit{k} would mean more anchor sizes to deal with in the RPN. Eventually we experimented with two sets of anchor sizes which were (9,26,70,208) and (8,18,38,88,248) corresponding to 4 clusters and 5 clusters respectively.

\subsection{Loss Functions}

The main purpose of the loss function experiments was dealing with the class imbalance. We used these losses in the final classification branch of the Faster R-CNN.

\subsubsection{Weighted Cross Entropy}

This loss only differs from standard cross entropy loss in the weightage given to individual classes. We compute this weight based on the number of samples of each class as follows:

\begin{equation}
  w_{c} = 1 - \frac{n_{c}}{\sum{n_{c}}}   
\end{equation}

where \(w_{c}\) is the class weight and \(n_{c}\) is the number of samples belonging to the class.

\subsubsection{Focal Loss}

The focal loss \cite{lin2017focal} is calculated as follows:

\begin{equation}
  p_{t} =
    \begin{cases}
      p & \text{if y=1}\\
      1-p & \text{otherwise}\\
    \end{cases}
\end{equation}

\begin{equation}
  \text{Focal Loss} = -(1-p_{t})^\gamma log(p_{t})
\end{equation}

Essentially this loss would give more weight to examples that the model finds difficult to classify (further the prediction from the true value, the greater the loss). 

\subsection{Data Augmentations}

First, it is important to note that for this project we work with 800x800 patches from the original dataset. In our experiments we have tried a total of three data augmentation strategies. All three of them include random horizontal flipping with a probability of 0.5. Apart from that the augmentations were as follows:

\begin{enumerate}
    \item Random resizing to one of (640,672,704,736,768,800) [\textbf{Augmentation 1}]
    \item Random resizing to one of (800,832,864,896,928,960) [\textbf{Augmentation 2}]
    \item Random cropping of a 400x400 region and resizing it to 800x800 during training. The idea behind this is to allow smaller objects to take up a larger part of the image. For inference, the whole image was resized to 1600x1600 to maintain the same object sizes in training and validating. [\textbf{Augmentation 3}]
\end{enumerate}

\subsection{Longer Pretraining}

We also tried using weights from differently pretrained models that were available. These models were Mask-RCNN models trained for a very long time (300 epochs, 1 epoch = 118000 COCO images) on the COCO dataset with Google's simple copy-paste augmentation and large-scale jitter as described in \cite{ghiasi2021simple}. We remove only the weights for the mask branch and use the rest as everything else is the same as Faster R-CNN. In pretraining, the networks were configured with batch normalization. We experimented with the weights with and without batch normalization.
\section{Results and Discussion}

The results of selected experiments on the validation set are summarized in Table \ref{tab:results} below. A qualitative example can also be seen in Figure \ref{fig:qual}. Other experiments were conducted that we believe added no value so they were not included. Our best model was a ResNext-152 model, achieving a 3.6 increase in AP over the baseline. Initially, changes were made one at a time to the experiments. Thereafter, changes that were found to produce good results were combined with later changes.

\begin{figure*}[h]
    \centering
    \includegraphics[width=0.7\linewidth, keepaspectratio=true]{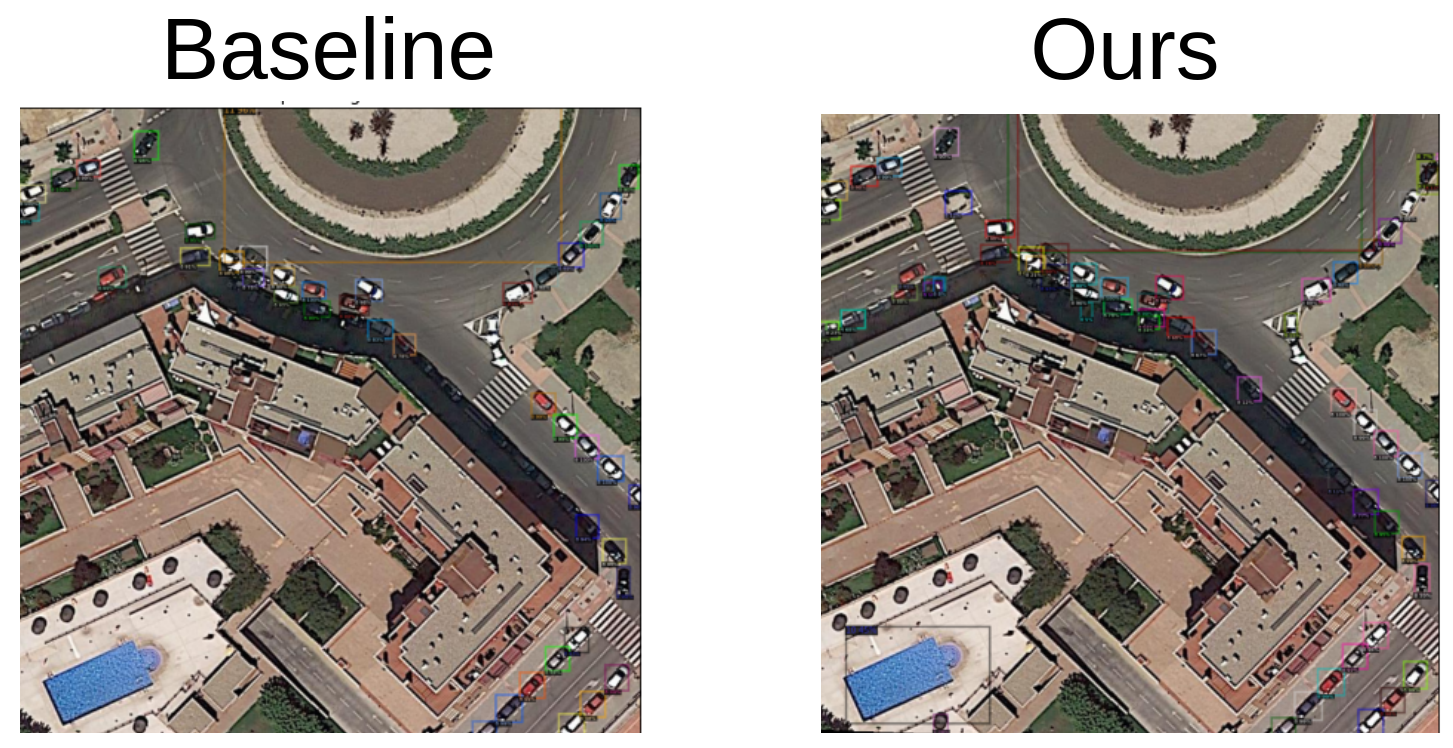}
    \caption{An example showing qualitative results on an example image from the dataset}
    \label{fig:qual}
\end{figure*}

\subsection{ResNet Tweaks}

Multidilation and changing the 7x7 conv to three 3x3s resulted in poor performance. We attribute this to the fact that these changes force us to discard the pretrained weights since the original layers are being changed. The benefit of pretraining was apparently significant and adds more value than tweaks in the architecture. Meanwhile, the last tweak was only a switch in terms of moving the strided convolution to the 3x3 conv in the middle of the bottleneck rather than keeping it in the first 1x1. Relatively, this was a small change and the pretraining weights were still used, not leading to a deterioration in performance. In fact, this was one of the better performing models and outperformed the baseline. The original structure of ResNet with the strided convolution being in the beginning of the bottleneck causes a big loss in information because the first layer is a pointwise convolution and the stride leads to many points being discarded completely.

\subsection{Backbone Changes}

Unsurprisingly, these were by far the most impactful changes to the architecture. While small jumps in AP were seen with other changes, completely replacing the backbone immediately resulted in a big jump in AP. This is expected because deeper, more advanced backbones provide much more information than shallow vanilla ResNets. In relation to other backbones, ResNext backbones were the best performing. Meanwhile, the Res2Net experiment was stopped early due to very poor performance in a short test run.

\subsection{Anchor Sizes}

We found anchors of sizes (16,32,64,128,256) to work best, outperforming the precomputed anchor sizes. This could possibly be due to suboptimal selection of other parameters such as the IoU threshold. Ideally, we could perhaps try different IoU thresholds with each of these but realistically this was difficult due to time and resource constraints. We also tried using all 5 anchors sizes for all feature maps rather than having a single size per FPN lateral output. This did not show a major improvement but we included it in our best model because there was no noticeable downside in keeping them.

\subsection{Loss Functions}

Results were poor for both the weighted cross entropy and the focal loss. A possible explanation for this is that adding these losses to the RPN rather than only the final classifier could have been a better alternative.

\subsection{Data Augmentations}

For the random resizing augmentations (Augmentation 1 and Augmentation 2) we found no significant difference in performance (about 0.1 AP). However, the random resized cropping was found to be harmful. We attribute this to the difference in input image size between training and validating. While we were aware of this problem, we would like to note that ideally, we would tile the validating images so they have the same size as the cropped training images. However, this would make the inference time comparisons unfair.

\subsection{Longer Training}

This was found to be helpful when tested with a ResNet-50 FPN backbone and perhaps could be done with a ResNet-101 FPN to outperform the current models in the future.

\begin{table*}[htb!]
\tiny
\begin{adjustbox}{width=1\textwidth}
\begin{tabular}{c c c c c c c c c c c c}
\hline
\textbf{Backbone} &\textbf{Modifications} &\textbf{Class Loss} &\textbf{Bbox Loss} &\textbf{Anchor Sizes} &\textbf{Anchor Rot} &\textbf{Aug} &\textbf{AP} &\textbf{APs} &\textbf{APm} &\textbf{APl} &\textbf{s/im}\\\hline
R-101 FPN &BASELINE &CE &Smooth L1 &32,64,128,256,512 &-90,0,90 &1 &37.146 &22.517 &44.185 &51.254 &0.02565 \\
X-152 FPN &- &CE &Smooth L1 &[16,32,64,127,256] &-90,0,90 &1 &41.8 &43.6 &53.9 &24.3 &0.12975 \\
X-101 FPN &- &CE &Smooth L1 &16,32,64,128,256 &-90,0,90 &1 &39.6 &41.5 &49.2 &18.3 &0.043 \\
R-101 FPN & Stride 2 in 3x3 &CE &Smooth L1 &[16,32,64,128,256] &-90,0,90 &1 &38.647 &22.41 &46.449 &56.403 &0.0295 \\
X-101 FPN &- &CE &Smooth L1 &32,64,128,256,512 &-90,0,90 &1 &38.034 &23.327 &45.223 &51.981 &0.044 \\
R-101 FPN &- &CE &Smooth L1 &16,32,64,128,256 &-90,0,90 &1 &39.9 &41.9 &49.6 &18.7 &0.025 \\
R-101 FPN &- &CE &Smooth L1 &16,32,64,128,256 &-90,0,90 &2 &37.807 &22.296 &44.628 &51.158 &0.0255 \\
R-50 FPN &Long Train &CE &Smooth L1 &[9,26,70,208] &-90,0,90 &1 &37.489 &22.252 &44.716 &52.844 &0.0225 \\
R-50 FPN &- &CE &Smooth L1 &[9,26,70,208] &-90,0,90 &1 &36.805 &22.705 &44.059 &49.975 &0.0193 \\
R-101 FPN &- &CE &Smooth L1 &8,16,32,64,128 &-90,0,90 &1 &36.56 &22.021 &42.16 &49.096 &0.026 \\
R-101 FPN &- &Weighted CE &Smooth L1 &32,64,128,256,512 &-90,0,90 &1 &35.796 &21.778 &41.495 &49.141 &0.0251 \\
R-101 FPN &7x7 conv to 3x3 &CE &Smooth L1 &32,64,128,256,512 &-90,0,90 &1 &31.96 &17.363 &37.245 &50.252 &0.026 \\
R-50 FPN &- &Focal &Smooth L1 &32,64,128,256,512 &-90,0,90 &1 &31.924 &19.603 &38.413 &40.469 &0.0199 \\
R-101 FPN &Multidilated &CE &Smooth L1 &32,64,128,256,512 &-90,0,90 &1 &29.038 &17.427 &35.687 &37.73 &0.0302 \\
R-101 FPN &- &CE &Smooth L1 &16,32,64,128,256 &-90,0,90 &3 &27.078 &12.269 &33.881 &46.148 &0.0255 \\
R2-101 FPN &- &CE &Smooth L1 &32,64,128,256,512 &-90,0,90 &1 &14.709 &10.333 &19.908 &8.928 &0.029 \\
\end{tabular}
\end{adjustbox}
\caption{Experimental results with different modifications on backbone, loss functions, augmentations, anchor sizes and angles during fine-tuning on the iSAID validation set}
\label{tab:results}
\end{table*}
\section{Conclusion}

As a result of all the experiments we have conducted, we concluded that while small modifications and tweaks can provide small increases in performance, the most impactful change is replacing the whole backbone. In the bigger picture, this also generalizes to the framework as a whole and not just the backbone of Faster R-CNN. More recent detectors such as YOLOv5 can provide much better results from the start, without modifications. This does not mean our experiments were exhaustive and there is no room for improvement but that is more practical path to take most of the time. It is also worth noting that in this project, this was not the path taken because the objective is to improve on the baseline architecture by modifying it rather than changing it completely.

{\small
\bibliographystyle{ieee_fullname}
\bibliography{egbib}

\begin{thebibliography}{10}\itemsep=-1pt

\bibitem{res2net}
Shanghua Gao, Ming{-}Ming Cheng, Kai Zhao, Xinyu Zhang, Ming{-}Hsuan Yang, and
  Philip H.~S. Torr.
\newblock Res2net: {A} new multi-scale backbone architecture.
\newblock {\em CoRR}, abs/1904.01169, 2019.

\bibitem{ghiasi2021simple}
Golnaz Ghiasi, Yin Cui, Aravind Srinivas, Rui Qian, Tsung-Yi Lin, Ekin~D Cubuk,
  Quoc~V Le, and Barret Zoph.
\newblock Simple copy-paste is a strong data augmentation method for instance
  segmentation.
\newblock In {\em Proceedings of the IEEE/CVF Conference on Computer Vision and
  Pattern Recognition}, pages 2918--2928, 2021.

\bibitem{he2017mask}
Kaiming He, Georgia Gkioxari, Piotr Doll{\'a}r, and Ross Girshick.
\newblock Mask r-cnn.
\newblock In {\em Proceedings of the IEEE international conference on computer
  vision}, pages 2961--2969, 2017.

\bibitem{resnet}
Kaiming He, Xiangyu Zhang, Shaoqing Ren, and Jian Sun.
\newblock Deep residual learning for image recognition.
\newblock {\em CoRR}, abs/1512.03385, 2015.

\bibitem{bagsOfTricks}
Tong He, Zhi Zhang, Hang Zhang, Zhongyue Zhang, Junyuan Xie, and Mu Li.
\newblock Bag of tricks for image classification with convolutional neural
  networks.
\newblock {\em CoRR}, abs/1812.01187, 2018.

\bibitem{fasterRCNN}
Jonathan Hui.
\newblock Understanding feature pyramid networks for object detection (fpn),
  Apr 2020.

\bibitem{kim2020lightweight}
Seung-Taek Kim and Hyo~Jong Lee.
\newblock Lightweight stacked hourglass network for human pose estimation.
\newblock {\em Applied Sciences}, 10(18):6497, 2020.

\bibitem{fpn}
Tsung{-}Yi Lin, Piotr Doll{\'{a}}r, Ross~B. Girshick, Kaiming He, Bharath
  Hariharan, and Serge~J. Belongie.
\newblock Feature pyramid networks for object detection.
\newblock {\em CoRR}, abs/1612.03144, 2016.

\bibitem{lin2017focal}
Tsung-Yi Lin, Priya Goyal, Ross Girshick, Kaiming He, and Piotr Doll{\'a}r.
\newblock Focal loss for dense object detection.
\newblock In {\em Proceedings of the IEEE international conference on computer
  vision}, pages 2980--2988, 2017.

\bibitem{redmon2017yolo9000}
Joseph Redmon and Ali Farhadi.
\newblock Yolo9000: better, faster, stronger.
\newblock In {\em Proceedings of the IEEE conference on computer vision and
  pattern recognition}, pages 7263--7271, 2017.

\bibitem{ren2015faster}
Shaoqing Ren, Kaiming He, Ross Girshick, and Jian Sun.
\newblock Faster r-cnn: Towards real-time object detection with region proposal
  networks.
\newblock {\em Advances in neural information processing systems}, 28:91--99,
  2015.

\bibitem{fasterRCNNPaper}
Shaoqing Ren, Kaiming He, Ross~B. Girshick, and Jian Sun.
\newblock Faster {R-CNN:} towards real-time object detection with region
  proposal networks.
\newblock {\em CoRR}, abs/1506.01497, 2015.

\bibitem{isaid}
Syed Waqas~Zamir, Aditya Arora, Akshita Gupta, Salman Khan, Guolei Sun, Fahad
  Shahbaz~Khan, Fan Zhu, Ling Shao, Gui-Song Xia, and Xiang Bai.
\newblock isaid: A large-scale dataset for instance segmentation in aerial
  images.
\newblock In {\em Proceedings of the IEEE Conference on Computer Vision and
  Pattern Recognition Workshops}, pages 28--37, 2019.

\bibitem{wu2019detectron2}
Yuxin Wu, Alexander Kirillov, Francisco Massa, Wan-Yen Lo, and Ross Girshick.
\newblock Detectron2.
\newblock \url{https://github.com/facebookresearch/detectron2}, 2019.

\bibitem{dota}
Gui-Song Xia, Xiang Bai, Jian Ding, Zhen Zhu, Serge Belongie, Jiebo Luo, Mihai
  Datcu, Marcello Pelillo, and Liangpei Zhang.
\newblock Dota: A large-scale dataset for object detection in aerial images.
\newblock In {\em The IEEE Conference on Computer Vision and Pattern
  Recognition (CVPR)}, June 2018.

\bibitem{resnext}
Saining Xie, Ross~B. Girshick, Piotr Doll{\'{a}}r, Zhuowen Tu, and Kaiming He.
\newblock Aggregated residual transformations for deep neural networks.
\newblock {\em CoRR}, abs/1611.05431, 2016.

\end{thebibliography}
}

\end{document}